\documentclass{article}
\usepackage{ijcai24}

\usepackage{times}
\usepackage{soul}
\usepackage{url}
\usepackage[hidelinks]{hyperref}
\usepackage[utf8]{inputenc}
\usepackage[small]{caption}
\usepackage{graphicx}
\usepackage{amsmath}
\usepackage{amsthm}
\usepackage{booktabs}
\usepackage{algorithm}
\usepackage{algorithmic}
\usepackage[switch]{lineno}
\usepackage{pdfpages}
\usepackage{graphicx}
\usepackage{wrapfig}
\usepackage{booktabs}
\usepackage{textcomp} 
\usepackage{soul}
\usepackage{amssymb}
\usepackage{adjustbox} 
\usepackage{siunitx} 
\title{Image Synthesis with Graph Conditioning: CLIP-Guided Diffusion Models for Scene Graphs
}

\author{
Rameshwar Mishra$^1$
\and
A V Subramanyam$^1$\\
\affiliations
$^1$Indraprastha Institute of Information Technology, Delhi\\
\emails
\{rameshwarm, subramanyam\}@iiitd.ac.in
}
\begin{document}

\maketitle
\begin{figure*}[h]
    \centering
    \includegraphics[width=1\textwidth,height=130pt]{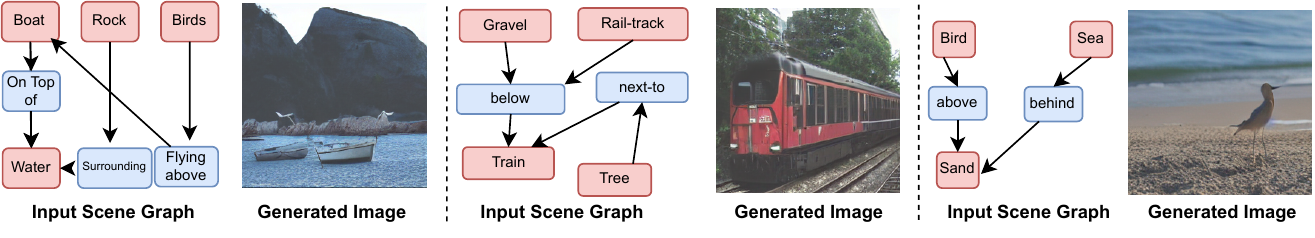}
    \caption{Samples generated using our architecture. It can be seen that the generated images reflect the structure given in the scene graph. They are different from ground truth images illustrating the diversity in generated samples. }
\end{figure*}
\begin{abstract}
    Advancements in generative models have sparked significant interest in generating images while adhering to specific structural guidelines. Scene graph to image generation is one such task of generating images which are consistent with the given scene graph. However, the complexity of visual scenes poses a challenge in accurately aligning objects based on specified relations within the scene graph. Existing methods approach this task by first predicting a scene layout and generating images from these layouts using adversarial training. In this work, we introduce a novel approach to generate images from scene graphs which eliminates the need of predicting intermediate layouts. We leverage pre-trained text-to-image diffusion models and CLIP guidance to translate graph knowledge into images. Towards this, we first pre-train our graph encoder to align graph features with CLIP features of corresponding images using a GAN based training. Further, we fuse the graph features with CLIP embedding of object labels present in the given scene graph to create a graph consistent CLIP guided conditioning signal. In the conditioning input, object embeddings provide coarse structure of the image and graph features provide structural alignment based on relationships among objects. Finally, we fine tune a pre-trained diffusion model with the graph consistent conditioning signal with reconstruction and CLIP alignment loss. Elaborate experiments reveal that our method outperforms existing methods on standard benchmarks of COCO-stuff and Visual Genome dataset.
\end{abstract}

\section{Introduction}

Scene graph represents a visual scene as a graph where nodes correspond
to objects and edges represent relationships or interactions between these objects. Improved generative models now allow users to generate high quality images where they can control the style, structure or layout of the synthesised images. Such conditional image generation allow users to guide the generation using text \cite{Rombach_2022_CVPR,pmlr-v139-ramesh21a}, segmentation mask \cite{Park_2019_CVPR}, class labels \cite{dhariwal2021diffusion}, scene layout \cite{jahn2021high,zheng2023layoutdiffusion}, sketches, stroke paintings \cite{meng2022sdedit}, and such more conditional signals. In particular, use of text as a conditioning modality offers a versatile approach, allowing for diverse combinations of inputs, encompassing intricate and abstract concepts. However, leveraging text for conditioning is not without challenges. Natural language sentences tend to be lengthy and loosely structured, relying heavily on syntax for semantic interpretation. The inherent ambiguity in language, where different sentences may convey the same concept, poses a risk of instability during training. This becomes particularly apparent in scenarios where precise description constraints are crucial. In this context, relying solely on text representations for a specific scene may prove to be insufficient.

Motivated by promising results of conditional generation and limitation of text as a conditional signal, in this work we propose a novel method to generate images from scene graphs. First introduced by \cite{johnson2018image}, scene graph to image generation is a task of generating images using set of semantic object labels and underlying semantic relationships among these objects. Most of the existing works follow a two stage architecture where they first generate a scene layout and use GAN to synthesize realistic images from these scene layouts \cite{johnson2018image,ashual2019specifying,ivgi2021scene}. Object nodes of scene graph is mapped to bounding boxes in the layout and the relationships are signified by the spatial structure of the layout. While these scene layouts can be effective in representing spatial relationships in the scenes, they fail to capture non-spatial complex relationships among objects. Relationships such as ``left of'', ``above'', ``sorrounding'' are spatial relationships, whereas ``looking at'', ``holding'', ``drinking'' are examples of non-spatial relationships. Translating scene graphs to accurate layouts and limiting representation capabilities of these layouts results in images inconsistent with the input scene graph. 

To overcome the limitations of existing methodologies, we propose to learn an intermediate graph representation while eliminating the need of predicting scene layouts. We use this graph representation as a conditional signal to fine-tune a pre-trained text-to-image diffusion model to generate images conditioned on scene graphs. To harness the strong semantic understanding offered by diffusion models, we propose to predict a conditional graph representation that aligns well with the inherent semantic knowledge of diffusion models. We use CLIP \cite{radford2021learning} guidance to generate such graph embeddings.

We first employ a GAN-based CLIP alignment module to train our graph encoder. This module instructs the graph encoder to generate graph embeddings that closely resemble the visual features of corresponding images in the CLIP latent space. To construct an effective conditioning signal for diffusion model, we fuse output of graph encoder with semantic label embedding of objects present in the scene graph. This conditioning signal is prepared to leverage the high prior semantic understanding of text-to-image diffusion models. We use this conditioning signal to fine-tune the diffusion model to generate images conditioned on scene graph.
We demonstrate the effectiveness of our method using established benchmarks like Visual Genome\cite{krishna2017visual} and COCO-stuff \cite{caesar2018coco}. Comparisons with current state-of-the-art methods reveal superior quantitative and qualitative results. We can summarise our contributions as follows:

(1) We propose to learn an effective graph representation, eliminating the need of predicting intermediate layouts to synthesize images. We use this graph representation to construct a suitable conditioning signal for text-to-image diffusion model. This conditioning signal is guided to leverage the semantic knowledge of text-to-image diffusion model.

(2) We propose a training strategy that effectively employs the constructed conditioning signal to fine-tune the diffusion model. 

Figure 1 shows the example images generated by our model with the input scene graph. Images generated by our model follow the input scene graph. For example, objects and relationships specified in the input scene graph in Figure 1, such as ``Birds flying above Boat'' and ``Sea behind sand'' are present in the generated images.
\section{Related Works}
\textbf{Diffusion as generative model.} The introduction of diffusion models by \cite{sohl2015deep} marked a notable approach to image generation. These models operate by learning the reverse process of the forward diffusion, where input is transformed into Gaussian noise. The denosing process is implemented using U-net \cite{oktay2018attention} or transformer \cite{vaswani2017attention} based models. In order to reduce the computation and training complexity, \cite{rombach2022high,vahdat2021score} introduced diffusion models which operate in latent space. \cite{dhariwal2021diffusion} proposed conditional generation by diffusion using classifier guidance. Recent advancements in these latent diffusion models have enabled users to produce diverse and realistic high quality images conditioned on various factors such as text \cite{pmlr-v139-ramesh21a}, artistic style, sketch, pose, and class labels \cite{esser2023structure}. Diffusion can also be used to generate text \cite{li2021bipartite,lin2023text}, videos \cite{esser2023structure,yu2023video} and graphs \cite{jo2022score}.  The conditioning is applied using cross attention mechanism between output of individual layers of denoising U-net and given conditional signals. Similar techniques are employed in text-to-image latent diffusion models. \cite{ramesh2022hierarchical} uses CLIP latent of text to condition high quality image generation. \cite{saharia2022photorealistic} uses latent from large language models such as T5 to condition latent diffusion models. While there has been notable progress in the diffusion-based conditional image synthesis, there exists a notable gap in the exploration of image generation from graph-structured data. 
In this work we explore the capabilities of text-to-image diffusion models in the task of image generation conditioned on scene graphs.

\textbf{Image generation from scene graphs.} Scene graph represents an image using set of nodes and edges. Nodes represent objects present in the image and their underlying relationships are captured by edges. Conventional scene graph to image methods tackle this task by following two stage architecture. At first a scene layout is predicted from graph. Scene layout represents an image using bounding boxes of corresponding objects present in the image. Scene layout is then translated into an image using convolution neural network based image synthesis models such as SPADE \cite{park2019SPADE}, OC-GAN \cite{sylvain2021object}. This task was first introduced by \cite{johnson2018image}. They employ a multi layer graph convolution network \cite{kipf2016semi} to get graph representation. This graph representation is used to predict object bounding boxes. The boxes are then used to generate images using cascaded refinement network. Generation is guided by GAN-based setup where a discriminator is employed to generate realistic images. Following \cite{johnson2018image} subsequent works adopt the two stage approach combined with GAN-based generation. \cite{ashual2019specifying} provides a way to control style of generated objects by providing a module to capture the style information of objects. \cite{herzig2020learning} use canonicalization for scene graph representation before translating it into scene layouts. This enhances the graph representation by incorporating supplementary information for semantic equivalence. \cite{ivgi2021scene} introduces an overlap loss to eliminate object overlapping. \cite{10.1016/j.cviu.2023.103721} use transformers for image generation. They learn layout representation using graph transformer. Further an image transformer coupled with VQ-VAE \cite{razavi2019generating} is used to sample images from these layouts. \cite{farshad2023scenegenie} uses scene layout and segmentation masks at sampling time of diffusion to generate graph aligned images. \cite{zhang2023learning} introduce a consistency module to overcome negligence of smaller object in the generated images. 

Most of the existing works utilise a layout based representation of graphs and GAN-based image generation. In this work we propose to use a graph representation which aligns well with the semantic prior of diffusion models. We use this aligned graph representation as a conditioning signal for diffusion based image generation. Notably, we eliminate the need for layout generation and convert the two stage to single stage generation.
\section{Method}
\begin{figure*}[t]
    \centering
    \includegraphics[width=\textwidth]{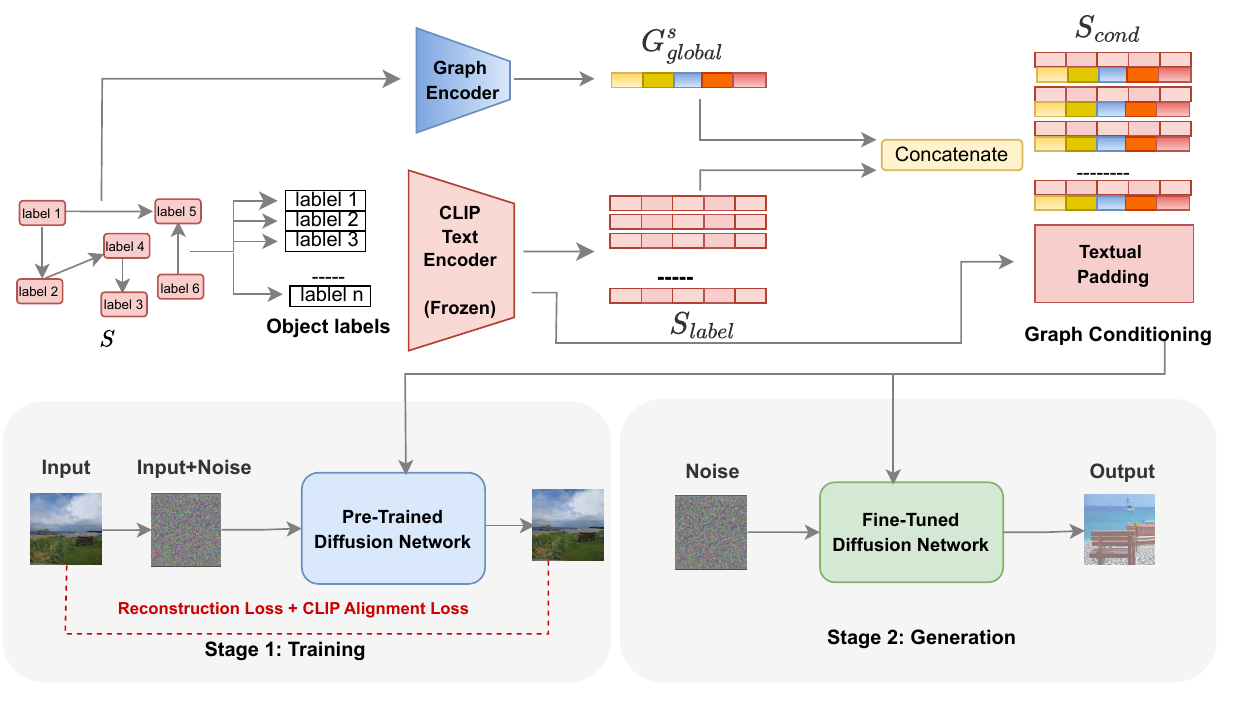}
    \caption{Overview of the proposed architecture. Graph encoder gives CLIP aligned graph embedding. This embedding is fused with semantic label embedding of objects present in the scene graph. The fused embedding forms a conditioning signal for diffusion model. During training, we pass this conditional signal with noise added input image and guide the training using reconstruction and CLIP alignment loss. During sampling we pass this conditioning signal with noise to generate image corresponding to the input scene graph.}
    \label{fig:main}
\end{figure*}
In this section, we present our proposed methodology with a detailed description of each component. We first give a brief overview of the conditional diffusion model.  Subsequently, we explain the functionality of the graph encoder, and the process of obtaining graph representations. In section \ref{sec:fine-tune}, we explain the creation of a CLIP-guided graph conditioning for diffusion model, along with the training strategy. Finally, we describe our GAN-based CLIP alignment module. An overview of our methodology is given in Figure \ref{fig:main}.

\subsection{Background for diffusion}
Diffusion models form a class of generative models designed to simulate the process of data generation through a series of diffusion steps. They constitute probabilistic generative models trained to comprehend data distributions through the sequential denoising of a variable sampled from a given distribution, mainly Gaussian. In the context of conditional generation, the goal is to generate data conditioned on some input information. For our case, we are concerned with pre-trained text-to-image diffusion model $\hat{x}_{\theta}$. Given noise $\epsilon \sim \mathcal{N}(0,I)$ and conditioning signal $S_{cond}$, the model generates an image $x_{gen}$ as follows: 
\begin{equation}
    {x_{gen}=\hat{x}_{\theta}(\epsilon, S_{cond})}.
\end{equation}
U-net based architecture is used to predict the added noise. The training is guided by a squared error loss to denoise an image or latent code with variable levels of noise. It is given by,
\begin{equation}
    {\mathbb{E}_{(x,\epsilon,t) } \lVert x - \hat{x}_{\theta} (x_t,S_{cond})\rVert_2^2},
\end{equation}
where $x$ is the reference image, $x_t=\alpha_t x + \sigma_t \epsilon$, is the noisy sample at diffusion time step $t$. $\alpha_t$ and $\sigma_t$ control the noise schedule. For latent diffusion models, latent embedding of input image is generated using VQGAN \cite{esser2021taming} or KL-Autoencoder \cite{rombach2022high}. All diffusion steps are applied in latent space, and then the final latent is decoded into an image.

\subsection{Graph Encoder} We use multi layer graph convolution network \cite{kipf2016semi,johnson2018image} to generate graph features from scene graph. We follow existing architecture of graph encoder for fair comparison with existing methodologies. Scene graph $S$ contains a set of objects $S_o$ and a set of relationships $S_r$. $S$ is represented using relationship triplets $(o_i,r_{ij},o_j)$ where $o_i \in S_o$ and $o_j \in S_o$ are two objects from object set $S_o$, and $r_{ij} \in S_r$ is the relationship between $i^{th}$ and $j^{th}$ object. Graph encoder fuses individual object embedding and individual relationship embedding to give a global scene graph embedding. For object $o_i$, we take a set $Out(o_i)$
to be the set of object to which $o_i$ has an outgoing directed edge. Set $In(o_i)$ denotes the set of objects where $o_i$ has an incoming directed edge from these objects. We find embedding for object $o_i$ as follows:
\begin{align*}
    G_{o_{out}}&= F_o^{out} (G_{o_i},G_{r_{ij}},G_{{o_j})_{j \in Out(o_i)}}, \\
    G_{o_{in}}&= F_o^{in} (G_{o_j},G_{r_{ji}},G_{o_i})_{j \in In(o_i)},\\
    G_{o_i}&=F^{pool}( (G_{o_{out}}) \cup (G_{o_{in}})), 
\end{align*}
where $G_{o_i},G_{o_j} \in R^{d_o}$ are the embeddings of object $o_i$ and $o_j$ respectively. $G_{r_{ij}},G_{r_{ji}} \in R^{d_r}$ are the embeddings for relationship $r_{ij}$ and $r_{ji}$ respectively. $F_o^{out},F_o^{in}$ are graph convolution layers and $F^{pool}$ is an average pooling layer. Similar to this, we find relationship embedding as follows:
\begin{equation}
    {G_{r_{ij}}=F^{rel} (G_{o_i},G_{r_{ij}},G_{{o_j})}},
\end{equation}
where $F^{rel} $ is a graph convolution layer.
After getting these individual object and relationship embedding, we calculate a global graph feature ${G_{global}^s}$ as follows. First we map each object and relationship embedding to same dimension $d_g$ 
using $F^{o}_{d_g}$ and $F^{r}_{d_g}$. $F^{o}_{d_g}$ and $F^{r}_{d_g}$ are 2 layer MLP's. They map $d_o$ dimensional object embedding and $d_r$ dimension relationship embedding to $d_g$ dimensional embedding respectively.
After getting same dimension embedding, we find an embedding to represent each individual triplet of scene graph as follows:
\begin{equation}
    {G_{triplet_{ij}}=F^{o}_{d_g}(G_{o_i})+F^{r}_{d_g}(G_{r_{ij}}) + F^{o}_{d_g}(G_{o_j}) }.
\end{equation}
Finally, a global embedding $G_{global}^s$, for a scene graph $S$ is calculated by concatenating individual triplet embedding and then mapping it to a $d_g$ dimensional feature.
\begin{equation}
    G_{global}^s=F^{global}_{d_g}(concat (G_{triplet_{ij}}))_{triplet_{ij} \in S}
    \label{eq:global_embed}.
\end{equation}
We use $G_{global}^s$ while creating a conditioning signal for training text-to-image diffusion model. 
\subsection{$G_{global}^s$ and CLIP Alignment}
\begin{figure}
    \centering
    \includegraphics[width=0.52\textwidth, keepaspectratio]{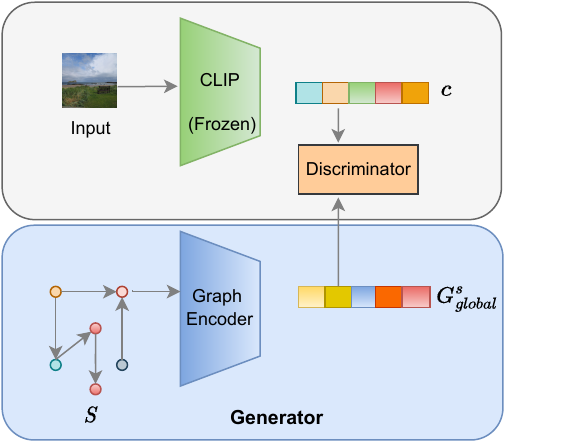}
    \caption{Graph embedding is aligned with CLIP visual features of the corresponding image. Alignment is achieved using GAN-based architecture.}
    \label{fig: alignment}
\end{figure}

We use GAN-based pre-training to align $G_{global}^s$ with CLIP features. Figure \ref{fig: alignment} provides an overview of our GAN-based CLIP alignment module. We consider graph encoder as our generator and $G_{global}^s$ as our generated data. CLIP visual features, $c$, form real data. Discriminator is trained to predict whether the input is from real or generated data. It guides the output of our generator to align with CLIP features. Training of graph encoder is guided by $\mathcal{L}_{\text{graph}}$, where $\mathcal{G}(S)$ is graph encoder output when given a scene graph $S$ as an input. $\mathcal{L}_{\text{graph}}$ is a standard generator loss for GAN-based architectures, ${\mathcal{L}_{\text{graph}} = \mathbb{E}_{S \sim p(S)} \left[ -\log \mathcal{D}(\mathcal{G}(S)) \right]}.$

\subsection{Training diffusion model with CLIP guided graph conditioning }
\label{sec:fine-tune}
$G_{global}^S$ captures the overall structure and interaction between the entities of scene-graph. However, this is global in nature. We hypothesize to use object label embeddings via CLIP with $G_{global}^S$ in our conditioning signal. The object encodings via CLIP can provide fine-grained semantic label details which can complement the global scene graph encodings. In Figure \ref{fig:main}, the semantic labels of objects present in the scene graph are passed through CLIP text encoder to generate object label embedding $S_{label}$. Then scene graph $S$ is passed through a graph encoder to generate embedding $G_{global}^s$ using Eq.\ref{eq:global_embed}. The object label embedding captures the entity level information of the image, while $G_{global}^S$ captures the interaction between these entities. Let $label_i$ be the semantic label of $i^{th}$ object present in $S$. Then, 
$${S_{label_{i}}=CLIPtext(label_i)}.$$
Finally we fuse $G_{global}^s$ and $S_{label_{i}}$ for all the labels to generate conditioning signal for fine-tuning the diffusion model.
$${S_{cond}=concat(G_{global}^s, S_{label_{i}}) \;\forall i \in S}.$$
We add textual padding to generate conditional signals of same dimension irrespective of the number of objects present in the scene graph. However, we note that pre-trained diffuion model we used is trained for text and image pairs. Thus it is necessary to design a training strategy to generate images conditioned on $S_{cond}$. Our training strategy is centered around optimizing the scene graph input for the diffusion model. This ensures that the scene graph aligns effectively with the input space of the diffusion model and weights of diffusion models are optimized for our conditioning signal. Experimentally, we verified that, when we simply pass our designed conditioning signal without fine-tuning the diffusion models, it results in the generation of low-quality images. These images lack coherence with the input scene graphs. Ablation results supporting the use of $S_{label}$ and the impact of fine-tuning are present in the supplementary material.

\subsubsection{Training objective}
Our learning objective is two-fold. First, diffusion model should learn the underlying distribution of image and scene-graph pairs. Second, we want to map output of graph encoder to a space where it aligns with prior semantic knowledge of text-to-image diffusion models. We achieve these goals in the following manner.


\textbf{Reconstruction Loss}: We use a reconstruction loss to guide the diffusion model to learn the underlying distribution of data. The loss $\mathcal{L}_{recon}$ is given by,
\begin{equation}
{\mathcal{L}_{recon}=\mathbb{E}_{(x,S) \sim {\text{data}}} \lVert x - \hat{x}_{\theta}(x_t,S_{cond}) \rVert^2_2},
\end{equation}
where $\hat{x}_{\theta}$ is denoising network of diffusion, $x_t$ is the noised sample at diffusion time step $t$, and $S_{cond}$ is the conditioning signal we curated. $(x,S)$ is the image-graph pair, sampled randomly from the data. 

\textbf{Alignment Loss}: Towards the second goal of aligning $G_{global}^s$ with CLIP space, we apply a mean squared error loss between $G_{global}^s$ and CLIP visual features of the corresponding image. 
For an image-graph pair $(x,S)$, the loss $\mathcal{L}_{CLIP}$ is given by, 
\begin{equation}
    {\mathcal{L}_{CLIP}=\mathbb{E}_{(x,S) \sim {\text{data}}} \lVert G_{global}^s - CLIP(x)\rVert_2^2}.
\end{equation}

Additionally, we also use Maximum Mean Discrepancy (MMD) loss \cite{li2017mmd} to bridge the domain gap between $G_{global}^s$ and CLIP visual features. In our experiments we observe that MMD loss makes the training stable, improves quantitative results and quality of images. MMD loss is defined as follows: 
\begin{align*}
    \mathcal{L}_{MMD}^2&=\mathbb{E}_{c \sim p({c})}\left[\phi(c,c)\right]\\
    &+ \mathbb{E}_{G \sim p(\text{graph})}\left[\phi(G_{global}^s,G_{global}^s)\right] \\
    &-2\mathbb{E}_{c \sim p({c}), G \sim p(\text{graph})}\left[\phi(c,G_{global}^s)\right],
\end{align*}
where $c$ is the CLIP feature of an input image and $G_{global}^s$ is the output of graph encoder for the corresponding scene graph. $\phi$ is the kernel function.
We combine both $\mathcal{L}_{CLIP}$ and $\mathcal{L}_{MMD}$ to define an alignment loss as,
\begin{equation}
    {\mathcal{L}_{align}=\beta \mathcal{L}_{CLIP} + (1-\beta) \mathcal{L}_{MMD}},
\end{equation}
where $\beta$ is a hyperparameter. 

\textbf{Total Loss}: Now by considering both reconstruction loss and alignment loss we define our training objective as follows:
\begin{equation}
    {\mathcal{L}_{train}=\lambda \mathcal{L}_{recon}+ (1-\lambda)\mathcal{L}_{align}},
\end{equation}
where $\lambda$ is a hyperparameter. We fine tune diffusion model and graph encoder using $\mathcal{L}_{train}$.

\renewcommand{\arraystretch}{1.2}
\begin{table*}[htbp]
\centering

\begin{tabular*}{\textwidth}{l | @{\extracolsep{\fill}} c *{3}{c}| c *{3}{c}}
\toprule
& \multicolumn{4}{c|}{\textbf{COCO-Stuff}} & \multicolumn{4}{c}{\textbf{Visual Genome}} \\

\textbf{Methods} (Reference) & {FID$\downarrow$} & {IS$\uparrow$} & {DS$\uparrow$} & {OOR$\uparrow$} & {FID$\downarrow$} & {IS$\uparrow$} & {DS$\uparrow$} & {OOR$\uparrow$} \\
\midrule
\textbf{SG2IM} (CVPR'18) $^*$ & 125.58 & 7.8 & 0.02 & {--} & 92.8 & 6.5 & 0.1 & {--} \\
\textbf{PasteGAN} (NeurIPS'19)$^*$ &70.2 & 11.28  & 0.60 & {--} & 130 & 6.5 & 0.38 & {--} \\
\textbf{Specifying} (ICCV'19)$^*$ & 68.27 & 15.2  & 0.67 & 70.84 &{--} &{--} & {--} & {--} \\
\textbf{Canonical} (ECCV'20)$^*$ & 64.65 & 14.5  & \ul{0.70} & 73.77 & 45.7 & 16.4 & \ul{0.68} & 72.83 \\
\textbf{RetrieveGAN} (ECCV'20) & 56.9 & 10.2  & 0.47 & {--} & 113.1 & 7.5 & 0.30 & {--} \\
\textbf{SCSM} (AAAI'22) & 51.6 & 15.2  & 0.63 & {--} & 63.7 & 10.8 & 0.59 & {--} \\
\textbf{SGTransformer} (CVIU'23)$^*$ & 52.8 & 15.8  & 0.57 & {--} &{50.16} &{14.6} & {0.59} & {--} \\
\textbf{SceneGenie} (ICCV'23) &63.27 & \ul{22.16}  & {--} & {--} & \ul{42.21} & \ul{20.25} & {--} & {--} \\
\textbf{LOCI} (IJCAI'23) &\ul{49.8} & 15.7  & 0.65 & \ul{81.26} & 44.9 & 14.6 & 0.62 & \ul{79.04} \\ 
\textbf{Ours}  &\textbf{38.12} & \textbf{30.18}  & \textbf{0.73} & \textbf{82.38} & \textbf{35.8} & \textbf{26.2} & \textbf{0.71} & {\textbf{81.04}} \\
\bottomrule
\end{tabular*}
\caption{Quantitative results on Visual Genome and COCO-Stuff dataset. All the results are either reproduced ($^*$) or taken directly from the original papers. Best results are shown in bold letters, and second best results are underlined. Results are reported for $256 \times 256 $ images.}
\end{table*}
\renewcommand{\arraystretch}{1}

\begin{equation}
    {\mathcal{L}_{\text{graph}} = \mathbb{E}_{S \sim p(S)} \left[ -\log \mathcal{D}(\mathcal{G}(S)) \right]}.
\end{equation}
Training of discriminator is guided by $\mathcal{L}_{\text{disc}}$, a standard discriminator loss for GAN-based architectures. Let $\mathcal{D}(c)$ be the predicted probability by discriminator when the input is from real distribution (CLIP features, $c$).
\begin{equation*}
    {\mathcal{L}_{\text{disc}} = -\mathbb{E}_{c \sim p(c)} \left[ \log \mathcal{D}(c) \right]- \mathbb{E}_{S \sim p(S)} \left[ \log(1 - \mathcal{D}(\mathcal{G}(S))) \right]}.
\end{equation*}
\begin{figure*}[t]
    \centering
    \includegraphics[width=\textwidth,height=380pt]{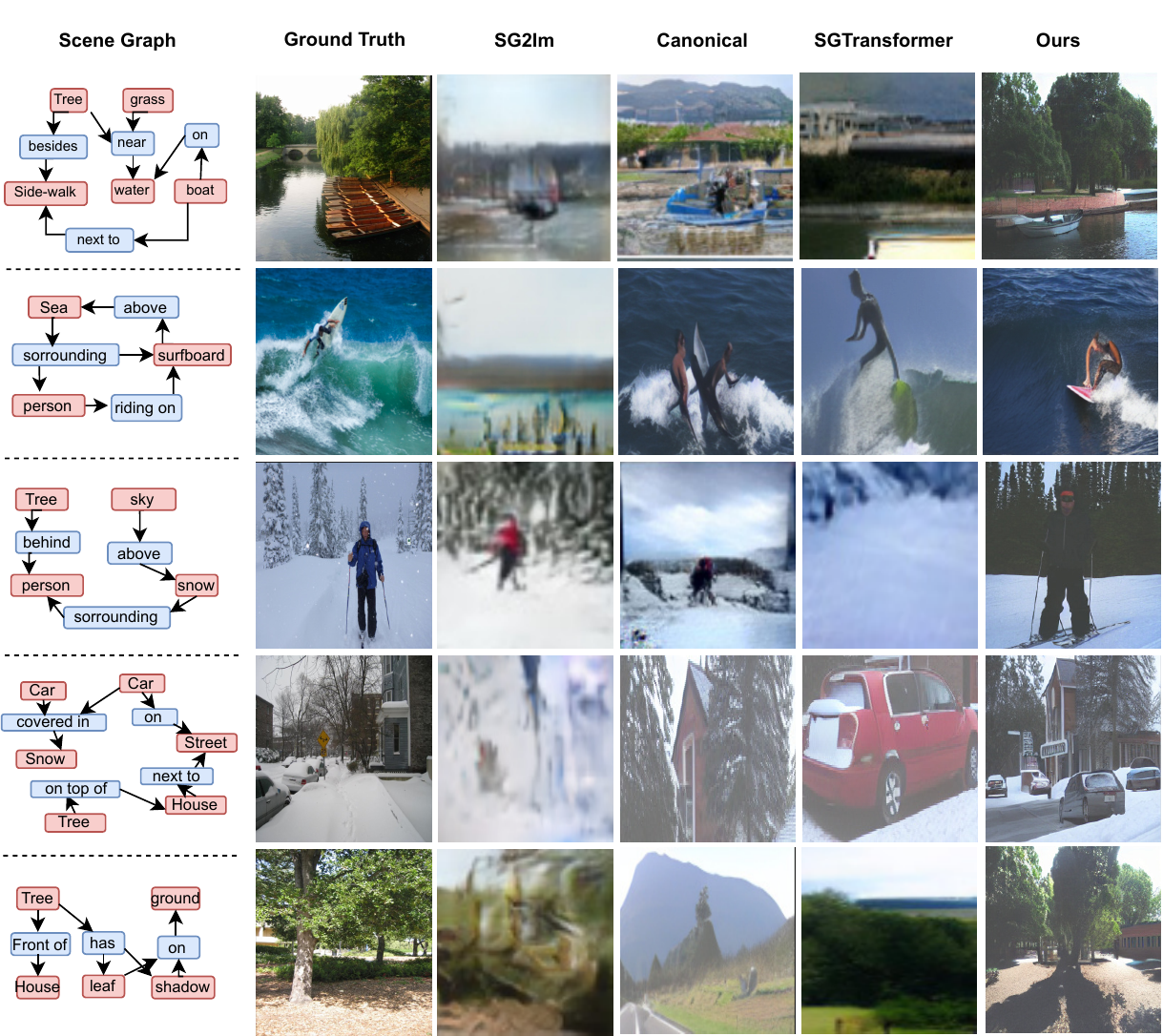}
    \caption{Qualitative comparison of $(256 \times 256)$ images generated by various publicly available scene graph to image models. All given input graphs corresponding to ground truth images are perturbed slightly to check effectiveness of each methods. The last columns shows images generated by our method. }
\end{figure*}
\subsection{Sampling process}
In Figure 3, stage 2 gives an overview for the sampling process. Once the diffusion network is trained, we can sample images from a latent noise $\epsilon$. For a fine-tuned denoising U-net $\hat{x}_{\theta}$, we can sample latent conditioned on scene graph $S$ as follows:
\begin{equation}
    {x_{latent}=\hat{x}_{\theta}(\epsilon,S_{cond})},
\end{equation}
where $S_{cond}$ is the curated graph conditioning signal and $\epsilon \sim \mathcal{N}(0,I)$. $x_{latent}$ is then decoded using diffusion's latent decoder to get an image. The generated image aligns well with the input scene graph.

\section{Experiments}
In this section, we outline the implementation details of our approach. We compare our results with existing state-of-the-art scene graph to image models. We verify effectiveness of each component of our training scheme by providing ablation results.
\subsection{Experimental setup}
\subsubsection{Dataset and Evaluation}
We train and evaluate our model on COCO-stuff and Visual genome dataset. We follow existing works \cite{johnson2018image,herzig2020learning} to filter out and divide the data into training and validation set for fair comparison. After pre-processing, we get 62,565 image-graph pair in training set and 5,506 image-graph pair in validation set of Visual Genome dataset. COCO-stuff has 40,000 and 5,000 image-graph pairs in training and validation sets respectively. We follow \cite{johnson2018image} to create synthetic scene graphs for COCO-stuff using spatial relationship edges.
To show effectiveness of our approach we evaluate our model using Inception Score (IS) \cite{salimans2016improved}, Fréchet Inception Distance (FID) \cite{heusel2017gans}, Diversity Score (DS) \cite{zhang2018improved}, and Object occurrence ratio (OOR) \cite{zhang2023learning}. IS is a metric commonly used to evaluate the quality and diversity of generated images in generative models. A higher Inception Score indicates better-performing generative models that produce both realistic and diverse images. DS is a measure used to quantify the variety and distinctiveness of generated samples for same input scene graph. FID evaluates the similarity between the distribution of real data and generated data using feature representations extracted from a pre-trained Inception model. OOR is the ratio of the objects detected in the generated image by YOLOv7 \cite{wang2023yolov7} with respect to the objects given in the input scene graph. High OOR implies high consistency of generated images with scene graphs.
\subsubsection{Implementation Details}
We use a pre-trained stable diffusion model \cite{rombach2022high}. Graph encoder is a standard multi layer graph convolution network taking nodes and edges as input. $d_g$ for graph encoder is 512, we take $\lambda=0.7$ and $\beta =0.5$. For reconstruction loss in diffusion, we guide are training with the MSE loss between predicted and added noise. We use Adam optimizer \cite{zhang2018improved} with a learning rate of 1e-6. We fine-tune the Diffusion model for 62,000 iteration and 32,525 iterations for Visual Genome and COCO-stuff datasets respectively,with batch size of 2. Discriminator is a 5 layer MLP, trained for 40 epochs with Adam optimizer.
\subsection{Results}
\subsubsection{Quantitative results}
Following previous works \cite{zhang2023learning,johnson2018image,herzig2020learning}, we have reported a comparison between our method and existing methods using FID score, IS, DS, OOR. Table 1 shows the effectiveness of our method based on these evaluation metric. 

On COCO-stuff benchmark, we are able to reduce FID score by 11.68, increase IS by 8.02 when compared to existing SOTA \cite{zhang2023learning}, \cite{farshad2023scenegenie} respectively. We also achieve the best results for DS score and OOR implying that the model generates diverse images yet contains the objects provided in the input scene graph. From Table 1 we can see that similar to COCO-stuff, there is significant improvement in terms of all the evaluation metrics for Visual Genome benchmark as well. Quantitative results show that we generate high quality and diverse images which are aligned with the given scene graph.
\subsubsection{Qualitative results}
Figure 4 show qualitative comparison between images generated by publicly available existing models and our method. We compare our results with SG2IM \cite{johnson2018image}, canonicalization based model \cite{herzig2020learning} and Transformer based modes SGTransformer \cite{10.1016/j.cviu.2023.103721}. Qualitative comparison shows superior performance of our model. It can be seen that generated images align well with the input scene graph and conserve the relationship structure provided by the scene graph. For example, in row 5 of Figure 4, images generated by canonical and SGTransformer contain trees, but fails to generate it's shadow. Similarly in row 4, SG2Im and canonical generate distorted images, whereas image generated by our model is most consistent with the input scene graph. More qualitative results are given in the supplementary material.
\subsection{Ablation study}
\begin{table}[h]
    \centering
    \begin{tabular}{lll}
        \hline
        Model type  & IS $\uparrow$ & FID $\downarrow$ \\
        \hline
        W/O GCA    & 27.6     & 43.28 \\
        W/O $\mathcal{L}_{align}$     & 28.72     & 41.17  \\
        W/O $\mathcal{L}_{MMD}$& 29.24     & 39.4 \\
        Ours ($\lambda=0.8,\beta=0.7$) & 29.74     & 39.28 \\
        Ours ($\lambda=0.6,\beta=0.3$)   &28.2  & 40.12 \\
        Ours & 30.18 & 38.12\\
        \hline
    \end{tabular}
    \caption{Results for ablation study done on COCO-stuff dataset. W/O is abbreviation for without. GCA refers to GAN based CLIP alignment module. $\lambda, \beta$ are hyperparameters used in training objective.}
    \label{tab:plain}
\end{table}
In this section, we illustrate the significance of each component in our training scheme. GAN-based CLIP alignment (GCA) module aligns output of the graph encoder with CLIP features of the corresponding image. This alignment is important since text-to-image diffusion models have strong semantic prior of CLIP features.
In Table 2, W/O GCA shows the performance without GCA. It is clearly evident that the incorporation of this module prior to fine-tuning the diffusion model results in improvements in both IS and FID scores.

For fine-tuning diffusion model, we introduce an alignment loss $\mathcal{L}_{align}$ with a standard reconstruction loss. After GAN-based alignment, this loss further guides the graph encoder to generate graph embeddings aligned with CLIP latent spaces.
We also take multiple combinations of hyperparameters $\lambda$ and $\beta$ to define our training loss $\mathcal{L}_{train}$. Finally, we show that our methodology containing GCA, $\mathcal{L}_{train}$ with $\lambda=0.7, \beta=0.5$ gives best results.
\section{Conclusion}
In this work, we propose a novel scene graph to image generation method. Our method eliminates the need of intermediate scene layouts for image synthesis. We use a pre-trained text-to-image model with CLIP guided graph conditioning signal to generate images conditioned on scene graph. We propose a GAN-based alignment module which aligns graph embedding with CLIP latent space to leverage the prior semantic understanding of text-to-image diffusion models. To further enhance the graph-conditioned generation, we introduce an alignment loss. Through comprehensive assessments using various metrics that measure the quality and diversity of generated images, our model showcases state-of-the-art performance in the task of scene graph to image generation.
\bibliographystyle{named}
\bibliography{ijcai24}

\end{document}


\maketitle
\section{Experimental Setup}
\textbf{Dataset.} We train and evaluate our model on COCO-stuff and Visual genome dataset. We process our data following existing works \cite{johnson2018image,herzig2020learning}.  After
pre-processing, we get 62,565 image-graph pair in training
set and 5,506 image-graph pair in validation set of Visual
Genome dataset. COCO-stuff has 40,000 and 5,000 image-graph pairs in training and validation sets respectively. We follow \cite{johnson2018image}  to create synthetic scene graphs
for COCO-stuff using spatial relationship edges.\\

\noindent{\textbf{Evaluation Metrics.}}  To show effectiveness of our approach we evaluate our model using Inception Score (IS) \cite{salimans2016improved}, Frechet Inception Distance (FID) \cite{heusel2017gans}, Diversity Score (DS)
\cite{zhang2018improved}, and Object occurrence ratio (OOR) \cite{zhang2023learning}. IS is a metric commonly used to evaluate the quality and diversity of generated images in generative models. A
higher Inception Score indicates better-performing generative
models that produce both realistic and diverse images. DS is
a measure used to quantify the variety and distinctiveness of
generated samples for same input scene graph. FID evaluates the similarity between the distribution of real data and generated data using feature representations extracted from a pre-trained Inception model. OOR is the ratio of the objects detected in the generated image by YOLOv7 \cite{wang2023yolov7} with respect to the objects given in the input scene graph. High OOR implies high consistency of generated images with scene graphs.\\

\noindent{\textbf{Training Parameters.}} We use a pre-trained stable diffusion model \cite{rombach2022high}. Graph encoder is a standard multi layer graph convolution network taking nodes and edges as input. $d_g$ for graph encoder is 512, we take $\lambda=0.7$ and $\beta =0.5$. For reconstruction loss in diffusion, we guide are training with the MSE loss between predicted and added noise. We use Adam optimizer \cite{zhang2018improved} with a learning rate of 1e-6. We fine-tune the Diffusion model for 62,000 iteration and 32,525 iterations for Visual Genome and COCO-stuff datasets respectively,with batch size of 2. Discriminator is a 5 layer MLP, trained for 40 epochs with Adam optimizer.

\section{Architectural Details}
\begin{wraptable}{r}{0.5\textwidth}
    \centering
    \begin{tabular}{l|l}
        \toprule
        Hyperparameter  & Considered value \\
        \midrule
        Input noise shape    &  $32\times 32 \times  4$ \\
        Noise scheduler      &  DDPM scheduler\\
        Diffusion timesteps & 1000     \\
        Autoencoder type & KL-regularized\\
        Learning rate scheduler & constant\\
        Unet's CA resolutions & 32,16,8\\
        \bottomrule
    \end{tabular}
    \vspace{5pt}
    \caption{Hyperparameter values for the diffusion model. Unet refers to the denoising network of diffusion. CA refers to cross-attention.}
    \label{tab:Table_1}
\end{wraptable}
Our architecture consists of three primary components: the GAN based CLIP alignment (GCA) module, a text-to-image diffusion model, and a graph encoder. This section provides architectural details for these components.
\subsection{Diffusion Network}
We use Stable Diffusion V1-4 checkpoint \cite{Rombach_2022_CVPR} as our diffusion model. The hyperparameter values for diffusion model is given in table \ref{tab:Table_1}. We employ the DDPM noise scheduler with 1000 diffusion timesteps. To generate 256$\times$256 images, we utilize an input noise latent of size 32$\times$32$\times$4.

\begin{table}[h]
    \centering
    \begin{tabular}{l|l|l}
        \toprule
        \multirow{2}{*}{Net.} & Layer(Input shape)  & Output Shape \\
                             &                & \\
        \midrule
        Object Net. & Linear (512)    & 512\\
          & ReLU      & 512\\
          & Linear (512)    & 512\\
          & ReLU      & 512\\
        \midrule
        Triplet Net. & Linear (3$\times$512)    & 512\\
          & ReLU      & 512\\
          & Linear (512)    & 512\\
          & ReLU      & 512\\
        \bottomrule
    \end{tabular}
    \vspace{10pt}
    \caption{Architecture of the graph convolution layer. Object network and triplet networks are joined Parallelly. All layers are sequentially added to create the respective network.}
    
    \label{tab:Table_2}
\end{table}
\begin{table}[h]
    \centering
    \begin{tabular}{l|l|l}
        \toprule
        \multirow{3}{*}{Net} & Layer(Input type/shape)  & Output shape \\
                             &                & \\
        \midrule
        Embedding Net. & Object Layer (Label)    & 512\\
          & Relation Layer(Label)      & 512\\
        \midrule
        Graph Net. & GraphConv (512,512$\times$3)    & 512,512\\
          & GraphConv (512,512$\times$3)    & 512,512\\
          & GraphConv (512,512$\times$3)    & 512,512\\
          & GraphConv (512,512$\times$3)    & 512,512\\
          & GraphConv (512,512$\times$3)    & 512,512\\
          \midrule
          Projection Net. & Avg Pool ($N_O\times512$)    & 512\\
          & Avg Pool ($N_T\times512$)    & 512\\
          & Linear (2$\times$512) & 512 \\
        \bottomrule
    \end{tabular}
    \vspace{10pt}
    \caption{Architecture of graph encoder Network. Object layer and relationship layer are two parallel embedding layers. Layers of Graph Network are sequentially connected. Projection net consists of two parallel average pooling layers. Output of these pooling layers is concatenated and fed to a linear layer.}
    
    \label{tab:Table_3}
\end{table}
\begin{figure*}[h]
    \centering
    \includegraphics[width=1\textwidth,height=280pt]{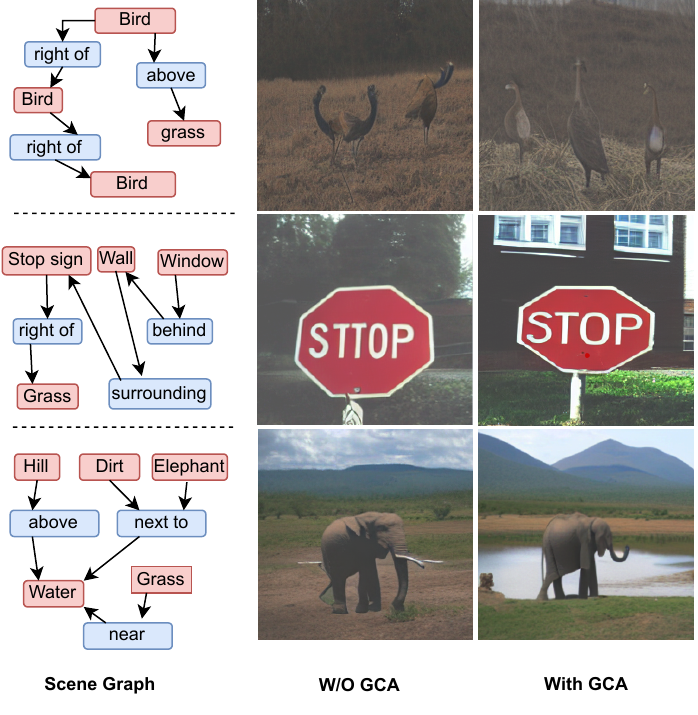}
    \caption{Qualitative results showing the effectiveness of GCA module. GCA refers to GAN based graph alignment. W/O is abbreviation for without. Column 1 contains input scene graphs, while Columns 2 and 3 display results generated without and with the use of GCA, respectively.}
    \label{fig: figure1}
\end{figure*}
\subsection{Graph Encoder}
Following previous works \cite{johnson2018image} we use a graph convolution network to encode our scene graph. Graph encoder consists of 5 graph convolution layers. Table \ref{tab:Table_2} shows the architecture of a single graph convolution layer. This layer consists of two parallel networks, one to predict object embedding and the other to predict triplet embedding. 

Table \ref{tab:Table_3} illustrates the comprehensive architecture of our graph encoder. Initially, object labels and relationship labels are fed into a vocabulary-based embedding layer. The input for the triplet network in the graph convolution layer is formed by concatenating the embeddings of the subject (S), relationship (R), and object (O) in a scene graph relationship triplet (S, R, O).  5 graph convolution layers are sequentially added to predict the object and triplet embeddings. In table \ref{tab:Table_3}, GraphConv takes two inputs, object embedding of size 512 and 512$\times$3 dimension concatenated input for triplet network. It outputs 512 dimension object and triplet embedding. We apply average pooling to get global object and triplet embedding. Finally, we project the concatenated global object embedding and triplet embedding to get our 512 dimension graph embedding.
\subsection{GAN based CLIP alignment module}
This module follows a standard GAN architecture. We consider graph encoder as our generator and it's architecute is given in table \ref{tab:Table_3}. Architecture of discriminator is given in the Table \ref{tab:Table_4}. We use clip-vit-base-patch32 checkpoint of CLIP to get CLIP features for GCA.
\begin{wraptable}{r}{0.5\textwidth}
    \setlength{\belowcaptionskip}{-35pt}
    \centering
    \begin{tabular}{l|l}
        \toprule
        Layer(Input shape)  & Output shape \\
        \midrule
        Linear (768)    &  256 \\
        BatchNorm      &  256\\
        LeakyReLU & 256     \\
        Dropout & 256 \\
        Linear (256)    &  128 \\
        BatchNorm      &  128\\
        LeakyReLU & 128     \\
        Dropout & 128 \\
         Linear (128)    &  1 \\
         Sigmoid (1) & 1\\
        \bottomrule
    \end{tabular}
    \vspace{5pt}
    \caption{All the layers are sequentially added to create the discriminator network. We use a negative slope of 0.2 for LeakyReLU. Dropout probability is 0.3}
    \label{tab:Table_4}
\end{wraptable}

\section{Additional results}
\textbf{Qualitative ablation results for GCA.} Figure \ref{fig: figure1} demonstrates that outputs generated with the use of GCA are more consistent with the input scene graph. For instance, in the first row, the model without GCA produces distorted birds, whereas in the second row, incorporating GCA leads to correctly spelled words in the image.\\

\noindent{\textbf{Additional qualitative results of our methodology versus existing approaches.}} Figure \ref{fig: figure2} showcases additional results, demonstrating the strong alignment of our method in generating images with the input scene graph. Our model produces diverse images. For instance, in row 3, both Canonical and SGTransformer generate outputs with a blue train structure, aligning with the ground truth containing a blue train. In contrast, our model generates an image featuring a red train. While our image maintains consistency with the input scene graph, it also introduces distinct elements, setting it apart from the original.

\begin{figure*}[t]
    \centering
    \includegraphics[width=1\textwidth,height=350pt]{images/supplementary_compressed.pdf}
    \caption{Sample images generated using different existing methods for comparison. It can be seen that our model generates high quality yet diverse images. Reference scene graphs are slightly perturbed to check effectiveness of each method.}
    \label{fig: figure2}
\end{figure*}
\newpage
\bibliography{references}